\documentclass{article}

\usepackage[english]{babel}

\usepackage[a4paper,top=2cm,bottom=2cm,left=3cm,right=3cm,marginparwidth=1.75cm]{geometry}

\usepackage{mathtools}
\usepackage{amssymb, amsthm}
\usepackage{amsfonts}
\usepackage{amsmath}
\usepackage{graphicx}
\usepackage[colorlinks=true, allcolors=blue]{hyperref}
\usepackage[percent]{overpic}
\usepackage[square,numbers]{natbib}
\usepackage{xcolor}

\newcounter{daggerfootnote}
\newcommand*{\daggerfootnote}[1]{%
    \setcounter{daggerfootnote}{\value{footnote}}%
    \renewcommand*{\thefootnote}{\fnsymbol{footnote}}%
    \footnote[2]{#1}%
    \setcounter{footnote}{\value{daggerfootnote}}%
    \renewcommand*{\thefootnote}{\arabic{footnote}}%
    }

\title{Protect Your Prompts:\\ Protocols for IP Protection in LLM Applications}
\author{M.A. van Wyk, M. Bekker, X.L. Richards, K.J. Nixon}

\begin{document}
\maketitle
\thanks{The authors are with the School of Electrical and Information Engineering, University of the Witwatersrand, Johannesburg, South Africa.}

\begin{abstract}
With the rapid adoption of AI in the form of large language models (LLMs), the potential value of carefully engineered prompts has become significant.\ \ However, to realize this potential, prompts should be tradable on an open market.\ \ Since prompts are, at present, generally economically non-excludable, by virtue of their nature as text, no general competitive market has yet been established.\ \ This note discusses two protocols intended to provide  protection of prompts, elevating their status as intellectual property, thus confirming the intellectual property rights of prompt engineers, and potentially supporting the flourishing of an open market for LLM prompts.
\end{abstract}

\section{Introduction}

LLMs, including those in the generative pre-trained transformer (GPT) family, are known to exhibit \emph{emergent properties} \cite{Bubeck2023}.\daggerfootnote{The term \emph{emergent property}, refers to the fact that a system exhibits novelties that are not due to the properties of any single part or subsystem of the system, but due to interactions among its subsystems \cite{RaineyHolland2022}.}\ \ Emergent behavior in such complex nonlinear adaptive systems manifests in a seemingly stochastic manner \cite{Bender2021}, which impacts directly on an LLM's responses to instructions for performing tasks given in the form of prompts.\ \ Consequently, querying an LLM repeatedly with the same prompt may yield different responses, while a tweak in a prompt may result in either no difference in an LLM's response, or a significant change.\ \ For critical applications, for example, in assistive surgery \cite{HeTangWang2023}, a substantial amount of time is spent on ensuring that the performance achieved is within an acceptable tolerance.\ \ Therefore, the monetary value of a well-crafted prompt (regardless of the field) which has painstakingly been developed through trial and error, including several hundred versions of iterative phrasing, and possibly also exploiting a particular LLM's architecture, will be considerable \cite{Kumar2023}.\ \ This has led to the emergence of a new field, called \emph{prompt engineering}, which refers to the art and science of engineering incantations that will evoke the desired response from an LLM \cite{WhiteFu2023,WangShi2023}.

This has underscored a simple fact that since the end of 2022, prompts themselves have become valuable.\ \ A prompt thus does not represent the desire of a user for an artifact that an LLM might produce, instead it stands as a proxy for the artifact it will ``unlock''.  Due to the inherent value situated in prompts, the risks of intellectual property (IP) and copyright violations have become real \cite{WangPan2023,PerezRibeiro2022}.  Protecting a prompt from being publicly viewable using some method, will enable prompt engineers (ranging from casual enthusiasts to large companies) to protect their competitive advantage in the market.  In this note we put forward suggestions on how prompts and associated IP protection may be accomplished.

\section{Terminology and Assumptions}
                                                                                
For the sake of brevity, we will refer to LLM prompts, i.e., the input or instruction sent to the model to generate a response, as ``task prompts'' or simply as ``prompts''.\ \ The prompt acts as a starting point or context for the model to generate a response.  Similarly, the generated output, as returned by an LLM, in response to a prompt, will be called a ``response'', and will be termed an ``artefact'' if it is deemed to have commercial value.\ \ We refer to the creator, coder, or engineer of a prompt as the ``prompt engineer'', and to any third party using the prompt as the ``user'', regardless of the distribution of IP-related rights. Therefore, the initial estimation of the commercial value of a prompt should be made by the prompt engineer.\ \ Finally, we use ``protocol'' to refer to a set of systematic guidelines that explain the ideal relations between different entities.

The protocols below assume that the targeted LLM is connected to the internet and permitted to access online content in real-time, i.e., beyond the online content on which it was trained.\ \ We assume that while the artifact related to a prompt is LLM-version contingent, the artifact can consistently and faithfully be returned, given the appropriate specifications contained within a prompt.

\section{The Problem of Prompt Protection}

In its present format of plain text, prompts are easily replicable and distributable.\ \ The utility of, and demand for, good prompts, and low barriers to entry into the domain, should have resulted in a lively and competitive market, with firms, scholars and enthusiasts representing a sizable proportion of the market participants.\ \ However, the monetary value of a well-designed prompt, the prompt engineer's competitive advantage, and the opportunity to profit financially from their creation, is undermined by the fact that once a prompt has been sold, it is effectively in the public domain, and open to the user to duplicate, modify, distribute, or resell it.\ \ This includes the possibility of a user undercutting the price, by selling it at a lower cost to potentially thousands of buyers.\ \ In theory, this contingency has had the effect of ostensibly driving the price for prompts down to zero, but in practice has prevented a market to be established at all.\ \ Prompts, as currently traded (or shared) are open to theft, and to alteration in nefarious ways, unintended by the original prompt engineer.\ \ Differently put, the value of a well-crafted prompt should be considerable, but is undermined by the problem of the lack of excludability.

Instead of producing prompts free of charge, the alternative approach a prompt engineer might take is to sell prompts for an unreasonably high price, i.e., for first access, which may compensate for the inherent opportunity cost associated with non-excludable goods.\ \ However, this would have a detrimental impact, choking what would otherwise have become a vibrant part of an LLM-related economy.  

These two considerations suggest that there would be significant value in approaches that lead to the protection of IP as they relate to prompts, in the service of a voluntary and competitive market for prompts.\ \ Below, we propose two protocols as ways to protect the intellectual property associated with a prompt.

\section{Proposed Prompt Protection Protocols}

In this section we suggest two protocols for protecting the IP of prompt engineers in increasing order of sophistication.

\subsection{Prompt Protection Protocol~1}

This is the most basic protocol of the two proposed, requiring minimal addition computational resources and hence minimal additional cost to be passed on to the end user.

\subsubsection{The Protection Mechanism}

\begin{figure}[h]
\hspace*{-7.5mm}
    \begin{overpic}[width=1.1\textwidth]{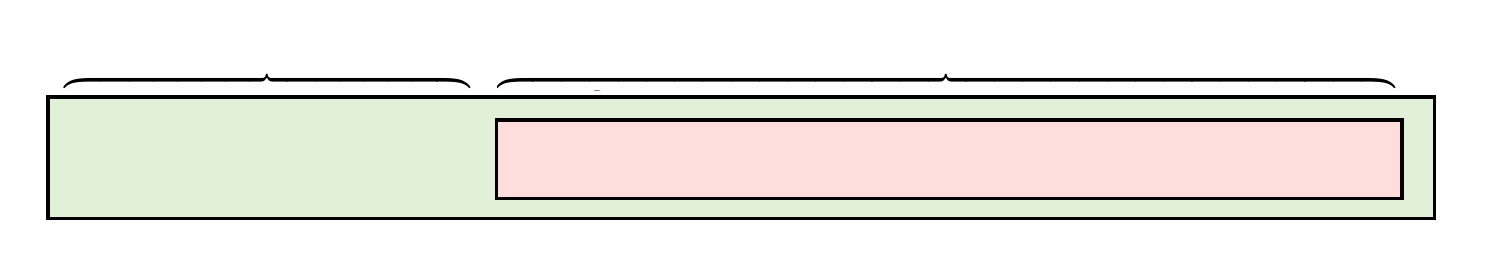}
        \put (8.5,14.5)   {Preamble \& Epilogue}
        \put (4,6.5)       {Decrypt, assimilate and forget:}
        \put (57,14.5)     {Task prompt}
        \put (35,6.5)      {$\zeta a\,\theta\ni\gamma$\textbackslash'$\Delta\in$$\neg\,\rho\,$\#$\,\bullet\,$\&$\,\notin$G$\times\!\sim$$\lambda\ntriangleright\,$@$\,\emptyset\,\mu\bowtie\dashv\,\equiv\!\nu$$\perp\!\exists\,\pi\,\eth\,\hbar\!\succ\ldots$}
    \end{overpic}
    \vspace*{-10mm}
    \caption{Suggested prompt composition for Protocol~1.} 
    \label{Protocol-1}
\end{figure}
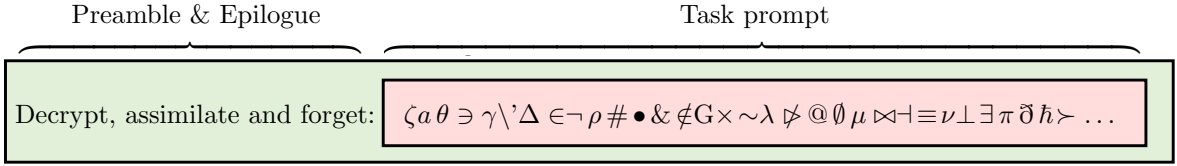

Here, the prompt prepared by the prompt engineer, consists of two parts concatenated into a single ASCII string, namely a human-legible preamble followed by a human-illegible core.\ \ The human illegible core is essentially the AI task prompt that forms the IP to be protected, and for this reason it exists in the form of an encrypted message.\ \ The composite prompt is depicted in Figure~\ref{Protocol-1}.\ \ In order for the user to unlock the potential of the AI prompt purchased, this encrypted message firstly has to be decrypted.\ \ This is the purpose of the legible preamble as it contains part of the information that is needed by the LLM to decrypt the task prompt.\ \ The remaining part of this information is stored elsewhere.\ \ Once decrypted, the LLM then processes the recovered task prompt which it was designed for, and upon completion of assimilating the task, the LLM is then required to completely forget the text describing the task (the prompt) to be performed.\ \ This requires the task prompt to possess an epilogue which explicitly instructs the LLM to forget the explicit task prompt.\ \ As a consequence, the LLM will have the ability to reason in accordance with, and execute, the task concisely described by the original task prompt, but will be unable to accidentally disclose the decrypted task prompt.

\subsubsection{Further Aspects on Protocol~1}

Protocol~1 requires a decryption bridge to be inserted between the user and the AI service provider.\ \ The preamble will be set up to contain decryption key which the user received from the prompt engineer (or their agent, such as a intermediary, who may be the IP owner) after having purchased the prompt.\ \ The decryption bridge may be located in one of three possible locations.\ \ The simplest option would be for it to be located on the user's server.  However, this poses a significant security vulnerability toward the IP owner particularly in the form of prompt interception \cite{WangPan2023}.

A better alternative, would be for the decryption bridge to be located on the IP owner's server.\ \ The user would then effectively communicate with the AI service provider via IP owner's server.\ \ However, spread out geographical locations of the three parties, e.g., over distant continents, could have a detrimental impact on the user's experience.  

The best alternative would be for the IP service provider to be allocated space on the AI service provider's server.\ \ The IP service provider's decryption bridge then intercepts all messages from the end user, decrypts these, and then send them to the AI server to be processed.\ \ This minimizes security risks as well as the adverse effects by avoiding having to route data through the IP owner's server.\ \ Although still exposed to the risk of poor network performance, this is the best that can be achieved.

As far as the decryption process itself is concerned, a broadcast or multicast decryption algorithm would be one option.\ \ Even though such algorithms allow for multiple decryption keys, the cost of these grow rapidly as the number of required keys increases.  A more viable alternative, though, would be to encode an issued user identification code together with the single decryption key into a virtually unique user key.  Upon receipt of the user key, the decryption bridge then extracts both the user identification code and decryption key.\ \ Only if the user key is valid, will the decryption bridge proceed to decrypt the encrypted message (the LLM task prompt), and only if the decryption key is correct will decryption be successful.\ \ Finally, the LLM then assimilates the instructions contained in the decrypted task prompt.\ \ Once the assimilation process of the task prompt has been completed, i.e., the artifact is produced, the task prompt will be deleted.\ \ The LLM is now able to respond to user instructions and queries, in the context of the assimilated task prompt, without any risk to the IP owner.  However, methods have been reported for extracting the original prompt from the LLM either by \cite{WangPan2023,PerezRibeiro2022} and consequently a robust forget strategy is essential for the success of this protocol.

\subsection{Prompt Protection Protocol~2}

Protocol~2 approaches the problem of IP protection by referring the LLM to a website which holds the key/cipher to the encrypted prompt.\ \ Under this model, prospective users purchase an encrypted prompt, based on a description of the resultant artifact, along with an instruction to the LLM regarding where to go to find the key.

\begin{figure}[h]
\centering
    \begin{overpic}[width=0.8\textwidth]{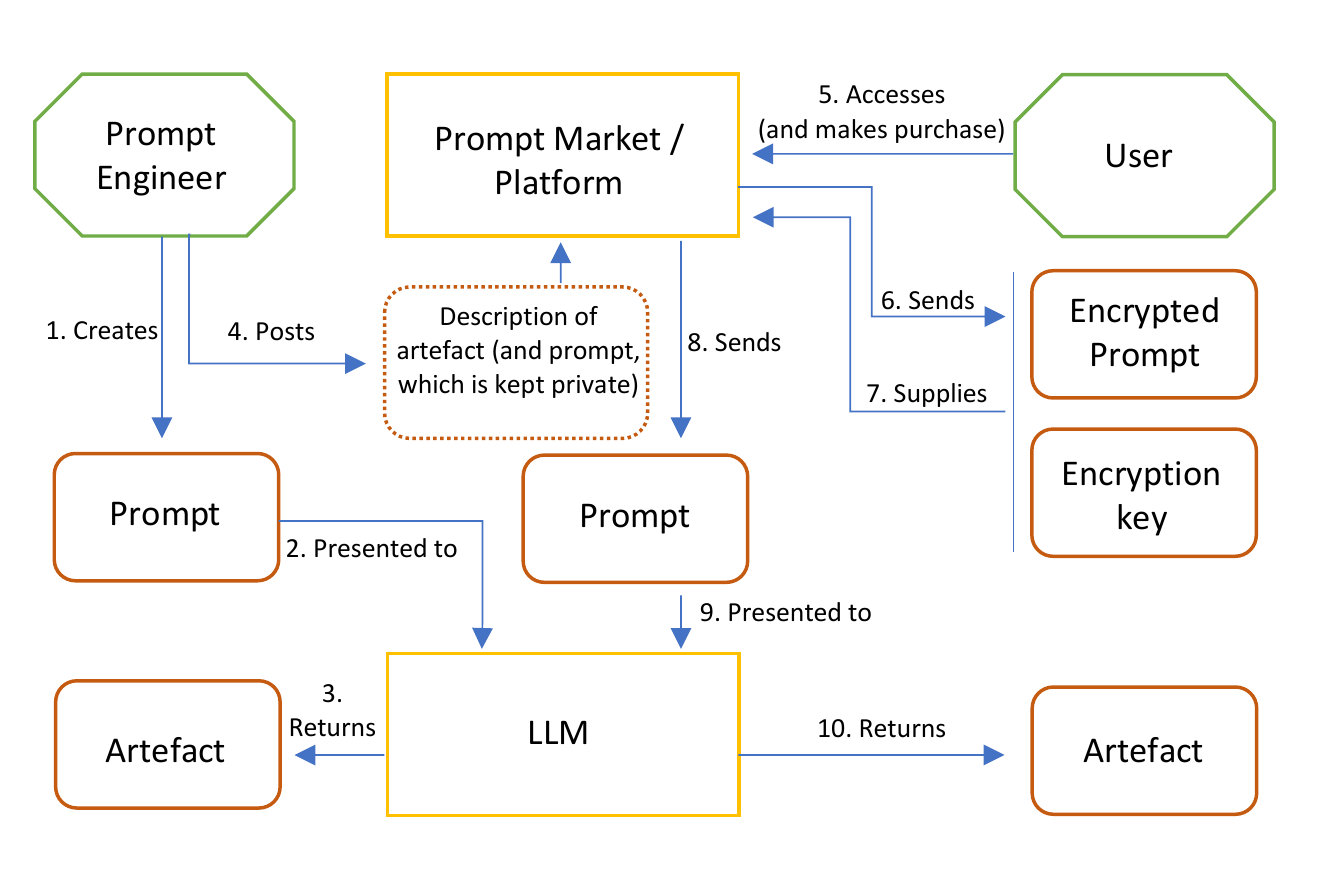}
    \end{overpic}
    \vspace*{-5mm}
    \caption{Schematic diagram of the keyed prompt process.  The numbered routes 1 to 10 represent the order of the process.} 
    \label{Protocol-2_Schematic}
\end{figure}

The key can thereby be located at a secure website that is accessed via an application processing interface, the details of which are included in the instruction.\ \ This arrangement allows the decryption-housing website to register a state-change once the key has been used, and alter the key, in order to prevent multiple calls on a single purchase, yet permitting a single prompt to be sold ad infinitum by the prompt engineer (or another agent with permission to do so).\ \ The schematic given in Figure~\ref{Protocol-2_Schematic} illustrates such a process.



\subsubsection{Further Aspects on Protocol~2}

A variation on this protocol is to host a platform that operates as a secure intermediary, where the prospective user is not required to access an LLM's website.\ \  Under such an arrangement, (encrypted) prompts are sold, accompanied by a single-use bearer token.\ \ The owner of a purchased prompt would access the intermediary platform (which may be the same site as where the purchase takes place), producing both the prompt and the token (which may be concatenated into a single string, or kept separately, allowing for two-step security).\ \ The token would verify whether the veracity of the purchase, and simultaneously trigger (1) a decryption, (2) an API call to the appropriate LLM, (3) returning the artifact and (4) invalidating the prompt for future use.

\section{Conclusion}

We have highlighted the increasing value of well-crafted LLM prompts.\ \ These prompts, carefully honed through iterative processes, have emerged as valuable tools that serve as proxies for the artifacts they generate.\ \ Their significance is evident in the growing demand for prompts and prompt engineering.\ \ To foster a thriving market for prompts, it is imperative to establish an open, transparent, and well-protected environment that encourages prompt engineers and users to engage in buying and selling activities, and which facilitates the responsible and ethical utilization of prompts across diverse domains.

However, a significant barrier to the establishment of such a market lies in the current non-excludability of prompts, and the absence of secure and exclusive mechanisms for their sale.\ \ This raises concerns about the protection of IP associated with prompts.\ \ Addressing this issue requires the development of protocols that can safeguard prompt-related IP, thereby enabling prompt creators to maintain ownership over their creations while ensuring fair access and utilization.

Within this note, we have explored two protocols for implementing IP protection for prompts across a wide range of applications.\ \ These protocols offer potential avenues for establishing a framework that balances the interests of prompt creators, users, and the broader market.\ \ By providing secure and exclusive mechanisms for prompt sale and ensuring the protection of prompt-related IP, these protocols lay the foundation for a vibrant and sustainable prompt market.\ \ Ultimately, the establishment of a protected prompt market will contribute to the advancement of LLM technology and its transformative impact on various fields.

Future work will focus on utilizing digital rights management (DRM) technology in the quest to protect IP and copyright associated with LLM prompts.

\bibliography{References}

\end{document}